# Clustering Concept Chains from Ordered Data without Path Descriptions

Kieran Greer, Distributed Computing Systems, Belfast, UK.

http://distributedcomputingsystems.co.uk

Version 1.2

**Abstract –** This paper describes a process for clustering concepts into chains from data presented randomly to an evaluating system. There are a number of rules or guidelines that help the system to determine more accurately what concepts belong to a particular chain and what ones do not, but it should be possible to write these in a generic way. This mechanism also uses a flat structure without any hierarchical path information, where the link between two concepts is made at the level of the concept itself. It does not require related metadata, but instead, a simple counting mechanism is used. Key to this is a count for both the concept itself and also the group or chain that it belongs to. To test the possible success of the mechanism, concept chain parts taken randomly from a larger ontology were presented to the system, but only at a depth of 2 concepts each time. That is – root concept plus a concept that it is linked to. The results show that this can still lead to very variable structures being formed and can also accommodate some level of randomness.

## 1   Introduction

This paper describes one attempt to cluster or create chains of higher-level concepts from individual lower-level ones. Because of its simplicity, the algorithm or mechanism relies on a number of very simple rules to determine what concepts belong together in a cluster. There were also some conditions on the data that was tested, but the results showed that the rules worked well, even over quite random looking structures and could be written in a generic or general way. The need to cluster lower-level concepts into higher-level ones has been looked at previously in the context of trying to automatically generate higher-level structures for reasoning over [1]. It is also now important as part of information systems however, especially with the emerging distributed sensor-based or mobile environments. In those cases, information that is related in some way might be received from different sources. As only partial information might be received, it would be helpful if it could be clustered or aggregated together, to give it more meaning. If the clustering mechanism is more generic, or does not require much supervision, then it is useful in many more scenarios.

The rest of this paper is structured as follows: section 2 describes the test environment that is used to evaluate the clustering accuracy. Section 3 describes some test results that verify the accuracy of the rules and the rules themselves. Finally, section 4 gives some conclusions on the work.

## 2   Test Environment

To test the clustering algorithm, data has been generated in the form of RDF [3] ontology scripts. The RDF ontology is hierarchical in nature, meaning that it is easy to give the concepts contained in it some sense of ordering. Each concept in the ontology is also unique, so a repeated concept in a result would be an indication that it might be incorrect. To make the test data interesting to read, it is based on behaviours. This also makes it easier to understand, where one example of an ontology is shown in Figure 1. This ontology describes the behaviour of two different people – John and Susan. This is presented to an evaluation system in the form of concept chains of length 2 each time. So, for example, the evaluating system might receive the information:

*jon_accessing, jon_reading:Book*.

This can be found under the John node in the ontology. The purpose of the testing is to link these concept parts together and evaluate them, to try and reconstruct the whole ontology. With 100% accurate information this is relatively easy to do and it is in fact simply an exercise in testing that the system works properly. If the system is always fed the correct information, it should simply be able to construct the correct links based on this. So the problem arises when the input data is slightly random or noisy in some way, when the system would need to determine what information is correct and what information is incorrect. This is where a few very simple rules can be applied to make the system more accurate under the following conditions:

- Only concept chains of length 2 are presented each time.

- The concepts in the chain have a definite ordering, so the second concept is a sub-concept of the first.

- Each concept in the ontology is unique, which means that there is only one correct position in the ontology for it. If the concept is repeated, then this indicates random data.

```
<rdf:RDF >
        <rdf:Description foaf:Person="John">
                <menow:jon_located>Bedroom</menow:jon_located>
                <menow:jon_moodString1>Sad</menow:jon_moodString1>
                <rdf:Description rdf:ID="jon_accessing">
                        <menow:jon_reading>Book</menow:jon_reading>
                </rdf:Description>
                <rdf:Description rdf:ID="jon_action1">
                        <jon_moveTo>Kitchen</jon_moveTo>
                        <rdf:Description rdf:ID="jon_action2">
                                <jon_cook>Cup of Coffee</jon_cook>
                                <menow:jon_moodString2>Happy</menow:jon_moodString2>
                        </rdf:Description>
                </rdf:Description>
        </rdf:Description>
        <rdf:Description foaf:Person="Susan">
                <menow:sus_located>Living Room</menow:sus_located>
                <menow:sus_moodString1>Bored</menow:sus_moodString1>
                <rdf:Description rdf:ID="sus_accessing">
                        <menow:sus_watching>TV</menow:sus_watching>
                </rdf:Description>
                <rdf:Description rdf:ID="sus_action1">
                        <sus_changeFocus>Music</sus_changeFocus>
                        <rdf:Description rdf:ID="sus_action2">
                                <sus_play>CD</sus_play>
                                <menow:sus_moodString2>Interested</menow:sus_moodString2>
                        </rdf:Description>
                </rdf:Description>
        </rdf:Description>
</rdf:RDF>
```

Figure 1. Example of a test ontology.

Previous work [1][2], has tried to link information using a path description. The path describes in more detail how the concepts are related to each other, giving the system the required level of accuracy to link them successfully. This was then tested by executing queries on the data and linking the results to provide for a more accurate search process. The linking mechanism also works in this case with no additional path information, but the threshold values need to be set manually. This paper looks at an alternative method, using a flat structure with some simple rules. The linking mechanism stored a value with each link that indicated how strong the link was. This value would be incremented when the link was used and then decremented when it was not used. In these tests, instead of using an increment and decrement, two values are stored for each potential concept of a chain (cluster). The first value indicates how many times the concept itself has been presented to be part of the chain. The second value indicates how many times any other concept in the chain has been presented as part of the chain. In a practical sense this has some advantages. There is no need to fine tune a weighting value, as only an increment is required each time. It is also clear the number of times each event has occurred, while if they are combined into a single value it might become less clear. You can, for example, just check whether the increment value for a concept itself is low, which would be an indicator that it is not valid. The results of the next section describe the clustering rules and how they work in practice. The ontologies were reconstructed after 300 presentations of chain parts to the evaluator.

## 3   Test Results

Two tests have been carried out. The first test presents 100% accurate information, to see what the results look like. The second test then introduces some random data to see if the clustering rules can tell what entries are incorrect. If the presented data is completely accurate, then the system would be expected to re-produce the ontology exactly. The only problem would be that enough concept chain parts are presented to cover the whole ontology. The advantages of storing the increment values both for the concept itself, and also for the chain as a whole, should be clear from Table 1. The first column of this table describes the concepts themselves. If the word 'chain' is included, then this is the base concept in the chain. If you look at the ontology of Figure 1 for example, the 'John' node contains 4 sub-nodes that represent the 4 concepts in its chain. The 'Own Inc' column represents the increment value for the concept itself when it is is presented to the chain, while the 'Chain Inc' column represents the increment value for when any other concept is presented to the chain, with this concept also present. If these two values are summed together, then it gives a more uniform value for the chain or cluster as a whole. There can be quite a large margin or difference, depending on when the concepts are included into the chain, but overall it helps to make clear what concepts actually belong to the chain and what ones do not. The summed value is shown in column 4. The two chains for 'John' and 'sus_action1' (first and last chains in the table), for example, show the differences.

| Chain or concept | Own Inc | Chain Inc | Sum |
|---|---|---|---|
| Chain: John | | | |
| jon_accessing | 26 | 66 | 92 |
| jon_action1 | 26 | 64 | 90 |
| jon_moodString1:Sad | 18 | 62 | 80 |
| jon_located:Bedroom | 22 | 56 | 78 |
| Chain: jon_accessing | | | |
| jon_reading:Book | 58 | 0 | 58 |
| Chain: sus_accessing | | | |
| sus_watching:TV | 56 | 0 | 56 |
| Chain: sus_action2 | | | |
| sus_moodString2:Interested | 66 | 46 | 112 |
| sus_play:CD | 46 | 60 | 106 |
| Chain: sus_action1 | | | |
| sus_action2 | 22 | 24 | 46 |
| sus_changeFocus:Music | 24 | 20 | 44 |

Table 1. Example of clustering information based on 100% accurate data.

Because the data should be clustered correctly with 100% accurate information, the system is tested with more random data as well. In that case, a concept chain part is created and then one of the concepts is replaced by a completely random one a specified number of times. For the next test, a correct concept was replaced by a completely random one, 1 time for every 10 chain parts that were created. If every chain part stores 2 concepts, then this is 1 out of every 20 concepts. While there are certain conditions to the data that is created and presented to the network, this can still produce quite random ontology structures, such as the one shown in Table 2. This table gives an example of the sort of data that can be presented and clustered based on the random information. The first 4 columns of this table are the same as for Table 1. The fifth column indicates if the entry is correct or not as specified by the real ontology structure. The sixth column describes the rule that would indicate an incorrect entry. A description of 'repeated' means that the own increment value is repeated somewhere else, while 'low value' means that the sum of the two increments is too low for the chain.

| Chain or concept | O - Inc | C - Inc | Sum | In Chain | Reason Not |
|---|---|---|---|---|---|
| Chain: John | | | | | |
|     jon_moodString1:Sad | 18 | 56 | 74 | ok | |
|     jon_action1 | 20 | 50 | 70 | ok | |
|     jon_located:Bedroom | 18 | 50 | 68 | ok | |
|     jon_accessing | 12 | 52 | 64 | ok | |
|     sus_action2 | 2 | 52 | 54 | x | repeated |
|     sus_watching:TV | 2 | 6 | 8 | x | low value |
|     jon_moveTo:Kitchen | 2 | 2 | 4 | x | low value |
| Chain: sus_action2 | | | | | |
|     sus_play:CD | 38 | 52 | 90 | ok | |
|     sus_moodString2:Interested | 44 | 44 | 88 | ok | |
|     jon_reading:Book | 2 | 82 | 84 | x | repeated |
|     jon_action2 | 2 | 54 | 56 | x | repeated |
|     sus_watching:TV | 2 | 12 | 14 | x | low value |
|     jon_moodString1:Sad | 2 | 4 | 6 | x | low value |

Table 2. Example of clustering information based on random data.

The 'John' chain, for example, has a lot of random concepts included that belong to other chains. If you sum the increment values for the concept itself and the other chain concepts, you get a correct value of around 60 – 70, shown in the 'Sum' column. This would include the top four concepts that actually make up the chain. You could then exclude the bottom two concepts because they have values that are too small. So this only leaves the concept 'sus_action2'. It was probably introduced early on and so it still has a relatively close overall total. However, the increment value for itself is only 2. This low figure suggests itself that it is a random concept and looking at other parts of the constructed ontology, the same concept occurs with the same random value for its own increment. Because the concepts should be unique and occur only once, this repetition of the smaller first value suggests that it should be removed from all chains where it has this value.

This therefore suggests some rules or guidelines as to whether a concept should be included as part of a chain, which can be summarised as follows:

Include an individual concept as part of a chain if:

- It has a sum total for its own increment and the chain increment that is close to other concepts in the chain.
- With the two increments, there can be some symmetry, for example a 20:80 value could also support an 80:20 value for another concept. This could be a key factor for including two counts.
- It has a unique own increment value and probably a larger own increment value to other occurrences.

Exclude an individual concept as part of a chain if:

- It has a sum total for its own increment and the chain increment that is appreciably smaller than the common value for the chain.
- It has an own increment value that is repeated in other chains.
- If its own increment value is not repeated exactly but is much smaller and close to other small values, then any larger value will be the correct one.

These rules have been applied to five test data constructions and shown to be accurate over all of the data. The following cases were the closest to invalidating one of the rules:

**Case 1:**

Chain: jon_moveTo:Kitchen

    sus_moodString2:Interested       2       2       4

| sus_accessing | 2 | 0 | 2 |
|---|---|---|---|

In this case, the 'jon_moveTo:Kitchen' concept does not actually have any sub-concepts, but the random data has added two of them. Looking at other chains would show that these entries have very small values overall, even if they are apparently consistent for this chain. Both concepts however are then repeated, which invalidates them because of their low counts.

**Case 2:**

Chain: jon_accessing

| jon_reading:Book | 24 | 10 | 34 |
|---|---|---|---|
| sus_changeFocus:Music | 6 | 26 | 32 |

In this case, the 'jon_reading:Book' concept is valid and the second concept 'sus_changeFocus:Music' has a sum total that is close to that one. Its individual increment value however is still appreciably smaller and it is repeated somewhere else with a similar increment value of 4. Applying the rules suggests that it is then invalid.

## 4   Conclusions

This paper has described a relatively simple counting mechanism that can be used to determine the correct concepts in a cluster, or higher-level concept chain. It works best however under certain conditions, which have also been described. If the incorrect information that is presented is completely random, then it is likely that this mechanism will be able to find the noisy entries and filter them out. If certain incorrect information is continually presented however, then the mechanism will probably take it to be correct, but possibly, also be able to notify the system about an inconsistency, through potentially more than one valid position in the ontology.

# 5   References

[1] Greer, K. Thinking Networks – the Large and Small of it: Autonomic and Reasoning Processes for Information Networks, currently published with *LuLu.com,* 2008, ISBN: 1440433275, EAN-13: 9781440433276.

[2] Greer, K., Baumgarten, M., Mulvenna, M., Nugent, C. and Curran, K. (2008). Stigmergic Linking for Optimising and Reasoning Over Information Networks, Research Report, found at http://licas.sourceforge.net/index_files/ReasonLink.pdf

[3] RDF (2004) RDF/XML Syntax Specification (Revised), W3C Recommendation 10 February 2004, found at http://www.w3.org/TR/2004/REC-rdf-syntax-grammar-20040210/#section-Syntax.